\documentclass[11pt]{article}

\usepackage[utf8]{inputenc}
\usepackage[T1]{fontenc}
\IfFileExists{lmodern.sty}{\usepackage{lmodern}\usepackage{microtype}}{\usepackage[protrusion=true,expansion=false]{microtype}}
\usepackage[english]{babel}

\usepackage[margin=1in]{geometry}

\usepackage{tabularx}
\usepackage{booktabs}
\usepackage{array}
\usepackage{caption}

\usepackage{indentfirst}
\usepackage[hidelinks]{hyperref}
\usepackage{xurl}
\usepackage{enumitem}
\setlist{topsep=0.2em,itemsep=0.2em}

\title{\textbf{Definitional alignment before capability alignment}\\[0.3em]
\large A Design-Science framework for adjudicating claims about AGI (DAF-AGI)}

\author{J. E. Aguilera Briones\\
\normalsize Postdoctoral Researcher, Administration and Business Innovation\\
\normalsize Universidad Internacional de Investigación México\\
\normalsize ORCID: \href{https://orcid.org/0009-0001-1121-1458}{0009-0001-1121-1458}}

\date{Preprint --- version 1.7. \quad Licensed CC BY 4.0.}

\begin{document}

\maketitle

\begin{abstract}
\noindent Claims that artificial general intelligence has already arrived and claims that it remains decades away are often defended from overlapping evidence. The disagreement is not exhausted by facts about systems: ``AGI'' lacks a single shared and stable referent and competing operationalizations can return different verdicts on the same system. This article treats that under-specification as a design and governance problem. Following Design Science Research Methodology, it develops DAF-AGI, a second-order conceptual artifact with two coupled components: five ordinal criteria for assessing the adjudicative fitness of candidate definitions and a structured governance audit of authorship, interest, certification, external verification and revision authority. The artifact is demonstrated on five prominent measurement families and one deflationary boundary position in a documented corpus and then stress-tested against a stylized strong arrival claim: that current generative systems constitute AGI because they outperform a well-educated adult on many cognitive tasks. The demonstration, conducted on evidence from the cited 2024--2025 sources, shows that the claim was certifiable only under a performance-based operationalization; capability-ontology, psychometric and skill-acquisition approaches did not certify it, the economic family remains indeterminate without labor-substitution evidence and the deflationary position refuses binary adjudication. The contribution is a novel integration and operationalization, not an empirical validation: corpus selection, scoring and demonstration were conducted by the author, while independent application, inter-rater testing and author-external cases remain necessary for validation. The paper further proposes \emph{definitional sovereignty} as an enabling component of algorithmic sovereignty: the institutional capacity to contest, certify and revise imported technological categories under public accountability.

\vspace{0.6em}
\end{abstract}

\vspace{0.5em}
\section{AGI has no shared referent and the vacancy is doing work}

A definition that returns ``yes'' and ``no'' to the same system,
depending on which operationalization governs, is not merely immature.
It is contested. The field of artificial general intelligence has spent
two decades treating its central term as if additional capability
evidence would eventually settle its meaning. Yet the literature has
long contained dozens of related but non-equivalent definitions of
intelligence and AGI {[}22{]}. The standard reading is that the science
is young. The reading defended here is narrower: the term sits over a
fault line of incompatible commitments and the absence of a shared
adjudication rule creates room for actors to advance thresholds aligned
with different scientific, commercial and regulatory purposes.

The evidence for incompatibility is no longer anecdotal. Three of the
most cited recent attempts to pin the term down disagree at the level of
what they measure. Google DeepMind's ``Levels of AGI'' separates depth
of performance from breadth of generality and places frontier language
models at ``Emerging AGI,'' a first rung, while reserving the label that
matches most prior conceptions of the term for a ``Competent AGI'' stage
no public system has reached {[}1{]}. A thirty-three author psychometric
proposal grounds AGI in Cattell-Horn-Carroll theory, defines it as
matching the cognitive versatility of a well-educated adult and on
applying its own battery finds a ``jagged'' profile: strong where
knowledge is dense, critically deficient in foundational machinery such
as long-term memory storage {[}2{]}. François Chollet's program rejects
performance as the unit of measurement altogether and defines
intelligence as the efficiency of acquiring skill on novel tasks,
operationalized through the ARC-AGI benchmarks where the best 2025
competition entry reached twenty-four percent on the private set of
problems that humans solve with ease {[}3{]}{[}4{]}{[}5{]}. These are
not three estimates of one quantity. They are three quantities.

The instability is not new and its history is instructive because the
field keeps relitigating the same fight under new names. Turing replaced
the unanswerable question of whether machines can think with an
operational substitute, the imitation game, precisely because the
underlying concept resisted definition {[}20{]}. A survey of the term's
usage two decades ago already collected more than seventy distinct
definitions of intelligence circulating in the technical literature and
grouped them only loosely {[}22{]}. When a research team claimed in 2023
to observe ``sparks'' of artificial general intelligence in an early
frontier model, the dispute that followed was not about what the model
did, which was largely agreed, but about whether what it did counted as
general intelligence, which was not {[}21{]}. Each episode follows the
same shape: capability advances, the advance is real and the community
discovers it has no shared rule for translating the advance into a
verdict about generality. The rule was never built. The term entered
wide circulation as a marketing and fundraising banner faster than it
was given testable content and the content has been supplied
retroactively, in incompatible ways, by parties with incompatible
interests.

The disagreement has a financial register. The chief executive of OpenAI
has publicly questioned the usefulness of the term while the company
continues to raise capital against the promise of building what the term
names. DeepMind, OpenAI
and Anthropic each circulate working definitions calibrated to their own
roadmaps and risk narratives. Expert surveys, meanwhile, place the
median arrival of high-level machine intelligence, defined as unaided
machines accomplishing every task better and more cheaply than human
workers, at 2047, with a ten percent probability by 2027 and full
automation of labor not reaching even-odds until 2116 {[}6{]}. When the organizations building the systems do not converge and when
their definitions carry different commercial and regulatory
implications, the indeterminacy cannot be treated as noise around a
single latent quantity. It is part of the object requiring analysis.

This diagnosis has a name and a lineage and naming it guards against
the charge of inventing a problem that the history of ideas already
described. Gallie's account of essentially contested concepts identified
a class of terms, drawn from politics, art and morality, whose correct
application inevitably and interminably divides competent users, not
because the users are confused but because the concept fuses description
with appraisal in a way no amount of evidence resolves {[}23{]}.
``Intelligence,'' and with it ``general intelligence,'' exhibits
several of the properties Gallie associated with such concepts: it is
appraisive, internally complex, open to
revision under new cases and applied by rival parties who each recognize
that the term is contested and use that contestation in their own favor.
Whether it satisfies Gallie's full set of conditions, in particular
mutual recognition of a shared exemplar among the contesting parties,
is not adjudicated here and the argument does not depend on it; what
the argument uses is the prediction the partial fit licenses, that the
disagreement will not close on additional capability evidence alone.
To treat AGI as exhibiting these properties is not to despair of
precision. It is to predict that the disagreement cannot be assumed to
close on capability evidence alone, to locate its source in the
appraisive core of the term rather than
in the immaturity of the field and to redirect effort from the search
for a uniquely true, interest-independent definition, which is
insufficient for governance even if one day a convention converges,
toward the governance of a currently
plural one. The value-laden critique of intelligence measurement reaches
the same conclusion from the side of practice, where benchmarks are
shown to encode contestable judgments about which capacities count
{[}7{]}. The two literatures support the more defensible claim this paper
operationalizes: persistent disagreement is not reducible to missing
capability evidence, because the candidate definitions encode different
units of achievement and evaluative commitments.

This produces a specific and underexamined kind of power. Whoever fixes
the operative definition of AGI fixes the moment at which ``we have
arrived'' becomes sayable and that moment is not cosmetic. It moves
valuations, triggers or defers regulation, reorders public research
priorities and rewrites the terms on which states and firms decide
whether to build, buy or depend. The capacity to set that criterion is
rarely named as a governable object. It is treated as something that
will be settled by progress. It cannot be assumed to be, because
capability progress does not by itself adjudicate between incompatible
value commitments
about what counts as intelligence in the first place {[}7{]}.

The contribution of this article is to make the vacancy operable.
Rather than propose another definition and argue for its truth, it
builds an instrument that takes a candidate definition as input,
exposes the commitments relevant to public adjudication and surfaces a
question technical comparisons often leave implicit: who is authorized
to adopt, certify and revise the operative criterion and who bears the
consequences of that choice. The instrument is named the
Definitional Alignment Framework for AGI, DAF-AGI and its central
output, the comparative matrix of five measurement families and one boundary position against
five scored criteria and a governance audit, is presented in Table~\ref{tab:daf-agi-matrix}
(Section 5.7). Its premise is a deliberate inversion of the field's vocabulary. The literature calls
``alignment'' the problem of making a capable machine conform to human
values. There is a prior definitional problem for the claims and
governance consequences whose operative condition is AGI itself:
arrival verdicts, regulatory triggers and contractual entitlements
cannot be adjudicated before the category they invoke is fixed and who
fixes it is presently informal. The scope of that priority is stated
exactly in Section 9; it does not extend to alignment research on
system behavior and risk, which proceeds without the label. The
community is debating how to govern AGI's arrival while it has not
aligned on what AGI is.

\section{What an adjudication artifact must do}

A useful instrument here is not one that ends the disagreement. It is
one that makes the disagreement legible, comparable and attributable.
Three requirements follow from that.

The first is \emph{procedural symmetry and explicitness}. The artifact
cannot be substantively neutral: its criteria embody a stated view of
what public adjudication requires. It can, however, apply those criteria
consistently to heterogeneous definitions, disclose the values embedded
in the procedure and allow users to contest or amend them. The
performance-based, economic, psychometric, skill-acquisition and
deflationary positions must therefore enter through the same documented
procedure even though they need not receive equal scores.

The second is exposure of commitments. Most definitions of AGI are
stated as if they were descriptions when they are in fact stipulations
carrying hidden parameters. ``Better than the average human at cognitive
tasks'' hides at least three: which humans, at which tasks, under what
conditions of access and effort. A definition's verdict is largely
determined by those buried parameters, so the artifact must drag them
into view and score them. The point is not to declare a definition
wrong. It is to show what it must assume to be right.

The third requirement is the one the literature systematically omits.
The artifact must treat the \emph{authority} behind a definition as a
variable, not a footnote. A criterion proposed by an entity that profits
from a particular verdict is not thereby false, but its independence is
compromised in a way that bears on its fitness as a public standard,
exactly as a safety standard authored by the firm it regulates is
suspect regardless of its technical content. Folding authorship, material interest, certification and revisability
into the evaluation is what fuses the analytical layer of the artifact
to its governance layer. Without that audit the framework would be one
more comparison table. With it, governance becomes an explicit and
contestable part of the output rather than background commentary.

These requirements define the artifact's success conditions in
design-science terms.

The requirements also explain why the comparison tables already present
in the literature are not sufficient. Several recent proposals survey
competing definitions and arrange them side by side and the better ones
are careful and informative. But a survey describes the disagreement; it
does not give a polity a procedure for governing it. In the comparison proposals examined in the selected corpus,
governance status is not treated as a structured output alongside
technical content. DAF-AGI is therefore distinguished neither by
claiming exhaustive coverage nor by discovering that definitions
differ, but by integrating a common adjudicative profile with an audit
of who authors, certifies and may revise the threshold. The intended
user is a definition-taker that must decide whether a candidate
criterion is fit to operate as a public standard.

\subsection{Contribution boundary and neighboring constructs}

What this paper does not claim to discover must be stated as precisely
as what it contributes, because several of its premises are established
results of adjacent literatures. That AGI is contested, that
definitions of intelligence encode values, that taxonomies have
political consequences and that benchmarks are not neutral are not
findings of this paper. The value-laden critique established them for
this domain and proposed contextual and participatory responses
{[}7{]}; the sociology of quantification established the general
mechanism by which measures bury judgment and remake what they measure
{[}24{]}{[}25{]}{[}26{]}; and the study of classification systems
established that categories are infrastructures of power whose
maintenance is political work {[}33{]}. The contribution claimed here
is narrower and specific: converting that shared diagnosis into a
second-order design artifact oriented to public adjudication, with
decision rules, a documented corpus and a structured institutional
audit of authorship, material interest, certification, external
verification and revisability, usable by a definition-taker rather than
only readable by a field insider.

The construct of definitional sovereignty requires the same
delimitation, because a reviewer could otherwise read it as a new label
for an existing combination. Its neighbors are real and the construct
borrows from several. The politics of classification {[}33{]} and the
commensuration literature {[}24{]}{[}25{]} supply the mechanism by
which a category exercises power; co-production accounts describe how
epistemic and social orders are made together {[}34{]}; the
standard-setting literature documents how private bodies come to author
rules with public force and how their committees are captured
{[}35{]} and rankings research describes how measures discipline the
measured {[}26{]}. None of these, however, names the specific institutional
capacity at issue here: the capacity of a definition-taking polity to
contest, certify against and accountably revise an imported threshold
for a contested technological category, as a precondition of governing
the systems classified under it. Classificatory power describes what
the category does; definitional sovereignty names what an institution
must be able to do about it and decomposes that capacity into
auditable components. If a reader judges that the construct reduces
without remainder to its neighbors, the audit and the artifact survive
the reduction, because their operation depends on the components, not
on the label. The construct's claim to a name rests on the gap it
covers between diagnosis and institutional capacity and that claim is
stated here so it can be tested rather than assumed.

DAF-AGI must function as a \emph{construct} (it introduces definitional
sovereignty as a named category), a \emph{model} (the structure of five scored
criteria and one governance audit) and a \emph{method} (the procedure for applying
that structure to any definition and to any claim built on one) {[}8{]}.
It is not and is not intended to be, an \emph{instantiation}: no
software, no automated scorer. This is a deliberate position within
design science, not a shortfall against it. The canonical taxonomy of
design-science outputs names constructs, models, methods and
instantiations as four distinct and equally legitimate artifact types
and an orthodox reading that treats only running instantiations as
``real'' artifacts misreads the tradition it invokes {[}8{]}{[}9{]}.
Theory-oriented and method-oriented design contributions are recognized
in their own right and a conceptual artifact is evaluated for the rigor
of its design and the soundness of its justificatory logic rather than
for a deployment it never promised {[}19{]}. The artifact here is
conceptual by design because its object, a contested definition, is
itself conceptual; an automated scorer would smuggle in the very value
choices the artifact exists to expose. Its evaluation is correspondingly
formative and artificial rather than naturalistic, a choice justified in
Section 7.

\section{Method: design science over a contested concept}

The methodological problem is unusual. The object of study is not a
system or a process but a word whose meaning is in dispute and the
deliverable is an instrument for governing that dispute. No single
method covers that terrain. The design is therefore a layered one, with
Design Science Research as the spine and three embedded techniques
supplying what the spine alone cannot.

Design Science Research is appropriate because the output is a designed
artifact addressing a wicked problem, a problem that is ill-structured,
value-laden and resistant to a single correct solution {[}9{]}. The
work follows the six activities of the Design Science Research
Methodology: problem identification and motivation, definition of
solution objectives, design and development, demonstration, evaluation
and communication {[}10{]}. Sections 1 and 2 discharge the first two
activities. Section 4 develops the artifact. Sections 5 and 6
demonstrate it. Section 7 evaluates it. The manuscript itself, released
as an open-access preprint, is the communication activity. This
sequencing is reported not as scaffolding but because traceability from
problem to artifact to evaluation is the property that distinguishes a
design contribution from an essay and that traceability is the standard
this paper claims to meet.

Three techniques are embedded in the design activity.

A \emph{documented definitional selection} builds the corpus to be
scored. The term is chosen with care: this is not a systematic review in
the PRISMA sense and it does not claim exhaustive coverage or a
registered protocol, because the artifact's demonstration needs a corpus
broad enough to exercise the five scored criteria and governance audit, not a census of every
published definition. What the selection does claim is traceability.
Definitions are drawn from peer-reviewed publications, recognized
technical reports and the stated positions of major laboratories; each
is recorded with its source, its proposing entity and the year;
near-duplicates are clustered into families by the unit of measurement
they adopt rather than by their wording. The clustering criterion
matters. Two definitions that both say ``human-level'' belong to
different families if one measures task output and the other measures
learning efficiency, because those commitments diverge under test. The
corpus reported in Section 5 is the result of that clustering and its
provenance is documented so that the scoring can be reproduced or
contested, which is the standard appropriate to a demonstration corpus
rather than the completeness standard appropriate to a systematic
review.

The selection rule is specified to make the corpus auditable rather than
impressionistic. Inclusion required that a definition be (i) stated with
enough precision to identify what it measures, (ii) attributable to a
named author, laboratory or standards body and (iii) either published in
a peer-reviewed venue, issued as a formal technical report, or adopted
as the operative position of an organization with material influence
over the field. Purely rhetorical invocations of the term with no
measurable content were excluded from scoring but retained as evidence
for the deflationary family of Section 5.6, because their emptiness is
itself a datum. The corpus is bounded in one further respect that matters for what follows: it collects definitions advanced as descriptive or measurement constructs, the objects conceptual engineering can score and sets aside the operative threshold frameworks that gate deployment or contractual rights by fiat, such as corporate safety levels and the contractual AGI triggers examined as an application in Sections 8 and 9, which the artifact reads through its governance lens rather than scoring on the five scored criteria and governance audit. Definitions were then assigned to families by a single
question applied uniformly: \emph{what does this definition count as the
unit of achievement?} The available answers, task performance, economic
substitution, position in a capability ontology, psychometric profile,
skill-acquisition efficiency and refusal of a unit, partition the corpus
for present purposes without claiming the partition is the only one
possible. The partition is by unit of achievement, an analytical choice made by the author rather than a natural kind discovered in the literature, but a choice anchored to a property each definition legibly exhibits in its own statement, so that the assignment is reconstructible by a second reader; its load-bearing role in the demonstration of Section 6 is examined there under a conceptual sensitivity check whose limits Section 6 states. A definition that combined units and several do, was assigned
to the family of its primary unit and flagged as hybrid, which matters
for the demonstration in Section 6, where the strong claim's persuasive
force is shown to depend on an unstated pairing of a permissive unit
with a reference-class label whose demanding operationalization the
claim never adopts.

\emph{Conceptual engineering} supplies the analytic operation at the
center of the artifact. The tradition descends from Carnap's notion of
explication, the deliberate replacement of a defective everyday concept
with a more precise successor fit for a purpose and has been developed
into an account of how concepts can be assessed and redesigned rather
than merely analyzed {[}11{]}{[}12{]}{[}13{]}. The relevance is direct.
``AGI'' is a defective concept in Carnap's sense: intuitive,
indispensable to current discourse and unfit as stated for the
adjudicative work the field demands of it. Conceptual engineering
licenses the move this paper makes, which is not to ask ``what does AGI
really mean'' but ``what would an AGI concept have to satisfy to do its
job and which candidates satisfy it.'' The five scored criteria and governance audit of Section 4
are the engineering specification.

The functionalist turn in this literature is what makes the move
legitimate rather than arbitrary. The question is not whether a
definition captures the essence of intelligence, a question the
value-laden critique shows has no interest-neutral answer, but whether a
definition is fit for a stated function. The function this paper
specifies is adjudication: a definition of AGI must be able to settle,
for independent competent evaluators, whether a given system has met
it, in a way that resists capture by the parties with a stake in the
verdict. Explication under that function has explicit desiderata in the
Carnapian tradition, similarity to the original concept, exactness,
fruitfulness and simplicity and the correspondence between those
desiderata and the artifact's structure should be shown rather than
asserted. Similarity is carried by C2 and C4: a candidate explicatum
must preserve the generality and the human comparison that make the
folk concept of general intelligence what it is, which is why a
definition satisfiable on a narrow, claimant-selected distribution
fails the desideratum however precise it is. Exactness is carried by
C1, C3 and C4: a stated test, an explicit position on autonomy and a
specified reference class are the three places where the corpus shows
adjudication-relevant vagueness concentrating. Fruitfulness is carried
by C5 together with the audit: a definition is fruitful for the
adjudication function when it can sustain comparable verdicts over
time, which requires both procedural stability and an accountable
locus of revision. Simplicity is carried by the artifact's two refusals,
no composite index and no scale finer than three levels, each adopted
because added structure would manufacture precision the evidence does
not contain. The governance layer then adds a
desideratum the classical tradition never needed: independence of the
explicator from the outcome. Conceptual engineering tells us how to
redesign a broken concept. The science-and-technology lens tells us why,
in this domain, the identity of the redesigner is part of the design.

The choice of adjudication as the governing function is itself a
contestable parameter and a framework that exists to surface buried
parameters must surface its own. At least three rival functions could
have been specified. A \emph{coordination} function would ask only that
a definition let heterogeneous actors converge on shared vocabulary,
which tolerates vagueness that adjudication cannot. A \emph{forecasting}
function would ask that a definition support calibrated prediction of
arrival, which rewards constructs tied to measurable trends regardless
of their fitness as public standards. A \emph{risk-gating} function
would ask that a definition trigger protective action early, which
rationally tolerates false positives that an adjudicative standard must
not. Adjudication is selected here not because the rivals are
illegitimate but because the consequences this paper is concerned with,
regulatory triggers, contractual rights, capital allocation and the
position of definition-taking states, are activated by binary public
verdicts of arrival and a verdict-issuing function is the one those
consequences presuppose. A reader whose purpose is coordination,
forecasting or risk-gating should expect the five scored criteria and
the governance audit to weight differently and the artifact's procedure, though not its present
scoring, transfers to those functions. Stating this converts the
deepest design choice from an assumption into an argued and revisable
position.

A \emph{science-and-technology-studies lens} governs the governance
layer. The premise that benchmarks and definitions of intelligence
encode values rather than discover them and that the encoding
distributes advantage, is taken from the value-laden critique of
intelligence measurement {[}7{]}. This lens is what permits the sixth
criterion to be a criterion at all. From a purely technical standpoint,
who proposes a definition is irrelevant to its content. From a
governance standpoint it is decisive, because a public standard derives
part of its legitimacy from the independence of its author. The artifact
treats AGI definitions as candidate public standards and so the lens is
not an ideological addition. It is entailed by the artifact's intended
use.

The integration is hierarchical, not eclectic. Design Science Research
provides the process and the artifact ontology. The definitional review
populates the demonstration. Conceptual engineering structures the
evaluation criteria. The science-and-technology lens justifies the
governance criterion and the definitional-sovereignty construct. Each
method occupies one slot. None is decorative.

\section{The artifact: DAF-AGI}

DAF-AGI has two components. The first scores definitions. The second
names and structures the authority that selects among them. The hinge
between them is the governance audit, which is why the framework can
claim to fuse analysis and governance rather than stapling one to the
other.

\subsection{Five scored criteria and one governance audit}

The criteria are selected and adapted for the stated function of
public adjudication. C1 through C5 draw on established traditions:
operationalizability on philosophy of science and measurement;
generality and autonomy on capability-ontology work; reference-class
specification on measurement theory; and stability on the requirement
that a public threshold remain interpretable across time. C6 is not an
ordinal measure of a definition's truth or quality. It is a governance
audit, adapted from work on standards, conflicts of interest and the
sociology of quantification. The contribution lies in integrating these
elements into one second-order procedure and applying them to AGI
definitions. The match in count between the five scored criteria and
the five measurement families is not a design constraint: the criteria
are fixed by the adjudication function, while the families are
partitioned by the units of achievement found in the selected corpus.

A candidate definition receives an ordinal profile on five epistemic
criteria: low, partial or high, each accompanied by a written rationale.
It then receives a separate structured governance reading. No composite
score is computed. Collapsing the profile and governance audit to a
single scalar would erase disagreements that the artifact is designed
to expose. Appendix A specifies decision rules sufficiently definite to
support a future inter-rater test; it does not claim that reproducibility
has already been established. All scores are relative to the
adjudication function specified in Section 3. A low score therefore
means that a definition serves that function poorly, not that it
misdescribes intelligence.

\textbf{C1 --- Operationalizability.} Can the definition be turned into
a test whose outcome competent parties would agree on? A definition that
cannot be failed cannot be passed and a criterion that supplies no
possible failure condition cannot support adjudication. This is a
requirement of operational determinacy, not of falsifiability:
definitions are not empirical hypotheses and no result refutes them;
what a candidate definition must supply is application conditions under
which competent evaluators can determine whether a given system
satisfies it.

\textbf{C2 --- Generality.} Does the definition make breadth across
materially different domains constitutive, through a domain structure
specified ex ante that cannot be altered after observing a candidate
system's performance without a declared revision, or through required
transfer to conditions the system was not prepared for? The criterion
is epistemic and deliberately silent on who authored the structure:
a proponent may fix a broad and rigorous taxonomy in advance and an
external source may impose a narrow or arbitrary one, so the identity
and interests of the author are recorded in the C6 audit, not here. This
is the criterion that the performance-superiority family
characteristically fails and it is the criterion that the ``G'' in AGI
nominally names. A definition satisfied by superiority on a task
distribution selectable or adjustable at the point of claim scores low
here however high the performance.

\textbf{C3 --- Explicitness on autonomy.} Does the definition state,
precisely enough for an adjudicator to apply, whether autonomous
goal-setting is required, excluded or graded? The criterion takes no
position on whether AGI must be autonomous. A definition that
explicitly declares prompted performance sufficient is as adjudicable
as one that explicitly requires autonomy; what fails the criterion is
silence, because an adjudicator confronting a silent definition cannot
know whether elicited performance counts.

\textbf{C4 --- Reference standard and comparison-class specification.}
What standard, baseline or comparison class determines achievement and
is it specified precisely enough to test? The standard need not be
human: an absolute capability threshold, a formal property or an
ecological adaptation condition can each anchor a definition and a
definition that is genuinely non-comparative is scored on the precision
of its stated standard, with the comparison-class component recorded as
not applicable. In the selected corpus every exemplar is human-relative,
so the criterion's work here is on the comparison class: ``better than
a human'' is unscorable until ``which human, at what'' is fixed and the
available classes are not interchangeable, since the median adult, the
well-educated adult, the domain expert and the entire human population
yield different thresholds. A definition that leaves its standard
implicit hides its own difficulty.

\textbf{C5 --- Procedural stability.} Does the same rule yield a
comparable interpretation over time? The criterion is temporal and
semantic, not institutional: it asks whether the threshold is specified
ex ante, whether changes preserve an audit trail, whether results
produced under earlier versions remain interpretable after revision and
whether revision applies prospectively rather than to results already
known. It deliberately does not ask who holds the authority to revise,
who approves a change, whether appeal exists or who represents the
affected, because those are facts about the distribution of
institutional authority and belong to the C6 audit. C5 evaluates the
continuity of the instrument; C6 records the authority over it. A
threshold that can be altered retrospectively after results are known
fails here whoever holds the pen.

\textbf{C6 --- Governance independence and accountability.} Who
authored the criterion, what material interests attach to its verdict,
who certifies achievement, what external verification or appeal exists,
and who may revise the criterion under what disclosure requirements?
The existence of an interest does not invalidate a definition and the
audit does not infer motive from benefit. It records institutional
conditions relevant to the definition's fitness as a public standard.

C1 through C5 assess a definition as an epistemic and measurement
object. C6 assesses the governance arrangement in which that definition
would operate. Both are applied within the same procedure, but they
produce different kinds of output. C6 is therefore neither an ordinal
score nor a proxy for truth. It is a structured reading across five
components: authorship, material interest, certification, external
verification or appeal and revision authority with disclosure: who may modify the rule, who approves the change, whether appeal exists and who represents the affected. This
difference in reporting format is intentional. Terms such as
``independence: partial'' would imply a precision the evidence cannot
support, while omitting governance would allow technical strength to
launder institutional capture.

\subsection{Definitional sovereignty}

The governance component formalizes what C6 audits into a positive
construct. \emph{Definitional sovereignty} is the institutional capacity
to set, contest and certify the operative criterion by which a contested
technological category is judged to be achieved, under criteria
generated or adopted through domestically accountable, plural and
contestable institutions rather than inherited from an interested
external author. The second clause is constitutive, not decorative,
because the construct would otherwise collapse into a weaker and more
dangerous one. A polity is not a unitary public interest: governments,
firms, research communities, labor, affected publics and security
apparatuses hold divergent stakes in where a definitional threshold
falls and transferring authorship from a foreign laboratory to a
domestic ministry or national champion relocates capture without
remedying it. Definitional sovereignty therefore has two faces that the
construct requires jointly: external autonomy, the capacity not to
inherit the criterion and internal legitimacy, the requirement that the
domestic criterion be publicly contestable, plurally representative,
transparent in its grounds, subject to administrative or judicial
review and guarded against domestic definitional capture by any single
sector. A state that seizes the definition for an unaccountable
domestic actor has changed the author of the threshold, not its
governance condition and the C6 audit applies to the domestic author
with exactly the force it applies to the foreign one.

The construct is built by analogy to algorithmic sovereignty,
understood as the institutional and strategic capacity to decide,
adapt, govern and audit algorithmic systems under one's own criteria of
public interest, data protection, security and operational continuity
{[}14{]}. Algorithmic sovereignty concerns control over deployed
systems; definitional sovereignty concerns the categories through which
those systems are classified and evaluated. The relation is enabling
rather than universally necessary: an actor may legitimately adopt a
shared or foreign-authored standard while retaining sovereignty if it
can audit the standard, participate in or contest its revision, certify
its local application and exit when the standard conflicts with public
interest. Dependency arises when these capacities are absent. Imported
definitions can therefore become durable governance infrastructure,
often persisting across changes of vendor or model, but they are neither
immutable nor intrinsically illegitimate.

The mechanism by which a definition becomes power is described by the
sociology of quantification and the construct inherits that mechanism
rather than asserting it. A benchmark or an index is a commensuration
device: it transforms heterogeneous qualities into positions on a common
scale and commensuration is a social process that buries the value
choices entering the scale while presenting its output as neutral fact
{[}24{]}. Quantification operates as a technology of objectivity that
travels well precisely because it strips away the visible marks of the
judgment that produced it, which is why numbers command trust from
parties who would contest the judgment if it were stated as a judgment
{[}25{]}. Definitional sovereignty is, in this vocabulary, sovereignty
over commensuration: the capacity to decide which qualities are rendered
onto the scale, by whom and with what acknowledged purpose. An entity
that authors the scale exercises power twice, once in the values it
encodes and once in the erasure of those values behind a number and a
definition-taker that adopts the scale inherits both the values and the
erasure. C6 exists to reverse the erasure. It re-attaches to a
definition the authorship and interest that commensuration conceals.

Definitional sovereignty has three observable components, which the
artifact uses to characterize a definition's governance status rather
than to score it numerically. The first is \emph{authorship}: who wrote
the criterion and against whose interest. The second is
\emph{certification}: who is empowered to declare the criterion met and
whether that authority is internal to the interested party. The third is
\emph{revisability under accountability}: whether the criterion can be
changed, by whom and with what public justification, which distinguishes
legitimate refinement from goalpost movement. A definition can be
technically strong on C1 through C5 and sovereign-deficient on all three
governance components and the artifact is designed to make that
combination visible rather than to let technical strength launder
governance capture.

The three components of the construct and the five components of the
C6 audit are related but not identical and the relation should be
stated rather than left to inference. The audit records five facts about
a particular definition: authorship, attached material interest,
certification authority, external verification or appeal and
revision authority with disclosure. The construct names the three
\emph{capacities} an institution needs in order to act on those facts:
to author or contest a criterion, to certify or refuse certification
against it and to revise it under its own accountability. Material
interest and external verification are thus inputs the audit documents,
while the construct specifies what a definition-taker must be able to
do about them. The audit describes a definition's governance condition;
the construct describes an institution's governance capacity.

The construct must be distinguished from its neighbors to avoid
absorbing their limits. \emph{Data sovereignty} concerns jurisdiction
over data; \emph{digital sovereignty} and \emph{technological
sovereignty} concern control over infrastructure, platforms and supply
chains. Each of these locates the contested object at a layer that can,
in principle, be acquired: data can be repatriated, infrastructure can
be built, suppliers can be diversified. Definitional sovereignty locates
the contested object one layer higher, at the categories against which
data, infrastructure and suppliers are evaluated and that object is not
acquired by procurement. A state can repatriate its data and still
regulate ``high-risk AI'' against a definition of risk authored abroad.
It can build domestic compute and still measure its ``algorithmic
maturity'' against an imported index. The category survives the
acquisition of every lower layer, which is why a sovereignty program
that stops below the definitional layer leaves its most durable
dependency untouched. The construct is not a synonym for the others. It
is the layer they omit.

\section{Five measurement families, one boundary position, one disagreement}

The documented definitional selection clusters the circulating
definitions of AGI into five measurement families by unit of achievement, plus one deflationary boundary position. The
clustering is by what each family counts, not by how it phrases the
count. The unit of scoring must be fixed before any score is read,
because families and definitions are different objects. Each family row
is an analytical ideal type in the Weberian sense, a construct built by
the author to make a unit of achievement comparable across cases, not a
definition any single actor advanced. The ideal type is anchored to a
named exemplar and C1 through C5 score that exemplar's operative
content, with the limits of generalization to co-members stated. The
C6 audit is never assigned to a family, because authorship,
certification and revision authority are properties of institutional
cases, not of ideal types: it is recorded per case in Appendix B and
the table's final column abridges the exemplar's case only. Where the
exemplar is itself a reconstruction, as in the performance family, the
record says so. Each family is then described, attributed and scored on
the five criteria, with profiles and rationale, consistent
with the decision not to compute composites.

\subsection{Performance superiority}

The performance-superiority family defines AGI as a system that matches
or exceeds human cognitive performance across tasks, often phrased as
exceeding human ability on ``any'' or ``most'' cognitive tasks. Public and popular formulations that treat broad human-level task
performance as sufficient belong here, including the intuitive position
that a system already better than the average person at writing, coding
and analysis has met the bar.

The scored exemplar of this family, the declared archetype, takes partial on C1: any completed instance is readily
measurable, but the definition itself fixes neither battery, aggregation
rule nor success condition, so a decision-relevant boundary remains
unspecified until a claimant supplies it, which is precisely what the
C1 rule marks as partial. It scores low on C2, because it imposes no
requirement of breadth beyond the chosen task set and is routinely
satisfied by systems that collapse outside their training distribution.
It scores low on C3, because it does not address autonomy: an
adjudicator cannot determine from the definition's text whether elicited
performance is meant to suffice, although in practice it is counted. Its C4 score is its decisive weakness, low, because
``human'' is left unspecified and the family's apparent success depends
entirely on quietly choosing a favorable reference class. On C5 it is
unstable, because the task set can be enlarged or narrowed after the
fact. Its C6 profile varies by application: task selection, certification and
revision may be decentralized, researcher-led or controlled by a system
provider. Where the claimant also selects the benchmark, the governance
risk is concentration of authorship and certification, not evidence that
the underlying performance result is false.

The performance family is where the strong claim examined in Section 6
lives and its profile already shows why that claim is contestable.

The family's strongest defense is pragmatic: if a system produces
expert-grade output across a wide span of paid cognitive work, the
insistence that it also satisfy some further criterion of generality
looks like metaphysics obstructing an obvious practical fact. The
defense is not negligible and it is the reason the family dominates
popular discourse. Its decisive weakness is equally simple. Performance
on a task set is only as general as the task set and the task set is
chosen by the party making the claim. A family whose verdict can be
engineered by curating the benchmark is a family whose apparent
generality is an artifact of selection and it is the one family in the
corpus that has never required a system to face a task its designers did
not anticipate.

\subsection{Economic substitution}

The economic family defines AGI by labor-market effect: a system, or
collection of systems, able to perform most or substantially all
economically valuable work at or above human level and more cheaply.
OpenAI's founding charter sits here, as do the
high-level-machine-intelligence and full-automation-of-labor constructs
used in expert surveys and the operational threshold, common in
forecasting and risk work, of automating more than ninety percent of
2025 economic roles on a feasibility rather than adoption basis
{[}6{]}{[}15{]}{[}16{]}.

The scored exemplar of this family, the Charter formulation, takes partial on C4. It names a concrete comparison class,
the human worker in an occupation, but leaves decision-relevant
boundaries around occupations, quality, supervision, cost and time. It
scores partial on C2: ``most economically valuable work'' spans domains
in intent, but the occupational set is bounded and selected at the
point of application, which is exactly what the C2 rule marks as
partial; intent cannot substitute for the decision rule.
Its C1 is partial: the criterion is measurable in principle but the
measurement is enormous and contested and ``feasibility not adoption''
is hard to settle without deployment. C3 is partial, because
substitution of a worker presupposes a position on autonomy that the
definition leaves recoverable from its structure but unstated. C5 is partial under the temporal rule: the labor benchmark moves
slowly, but no versioned trail fixes how the occupational baseline
would be reinterpreted across time. On C6 the family is mixed: it
underwrites both the deferring posture, useful for those minimizing
regulatory exposure and the arriving posture, useful for those raising
capital and the same firm can occupy both depending on audience.

The family's strongest defense is that it ties an abstract debate to a
measurable social consequence: whatever AGI is, its arrival matters
because of what it does to work and a definition pegged to labor
substitution measures the thing society actually cares about. The
criticism is that economic value is not a cognitive property. A system
can displace a worker for reasons of cost and scale that have nothing to
do with matching human cognition and a system can match human cognition
in domains with no labor market attached. Defining general intelligence by market effect therefore combines a
capability claim with deployment economics. That may be appropriate for
a labor-impact forecast, but it should not be presented as a direct
measurement of cognition without argument.

\subsection{Capability ontology}

The capability-ontology family does not offer a single threshold. It
offers a structured space and locates systems within it. DeepMind's
``Levels of AGI'' is the exemplar: it separates performance (depth) from
generality (breadth), adds autonomy as a third axis and defines
ascending levels from emerging through competent, expert, virtuoso to
superhuman, each applicable to narrow or general systems {[}1{]}.

The scored exemplar of this family takes partial on C1: it is designed to be operationalized
through benchmarks, but its own authors present requirements that
future benchmark suites would have to satisfy rather than an executable
battery, so the test exists in principle while its execution remains
unspecified at decision-relevant points, the condition the C1 rule
marks as partial. It scores high on C2, because generality is an explicit
axis rather than an assumption. It scores high on C3 under the
explicitness rule: the framework states its position on autonomy
precisely, grading it as a separate dimension of human-system
interaction rather than a condition of the performance-generality
classification, so an adjudicator knows exactly what role autonomy
plays. C4 is partial because
levels are pegged to human percentiles but the percentile-to-level
mapping requires further specification. C5 is partial: distributing the
question across levels resists a binary goalpost move and the framework
is publicly versionable, but no institutionalized procedure fixes how
levels or benchmark choices would be revised, which leaves revision
discretionary at a decision-relevant point. Its C6 audit records that the framework is laboratory-authored and
publicly documented, while benchmark selection, certification and future
revision remain linked to the authoring institution. This is a
governance fact, not evidence that its ``Emerging'' placement is
strategically motivated.

The exemplar combines explicit architecture with unfinished
operationalization, a profile the recalibrated scoring now records
rather than smooths over and it illustrates why C6 must remain
separate: epistemic strength and governance status are settled by
different evidence.

\subsection{Psychometric parity}

The psychometric family imports the most empirically validated model of
human cognitive abilities, Cattell-Horn-Carroll theory and defines AGI as
matching the cognitive versatility and proficiency of a well-educated
adult across the model's core ability domains, operationalized by
adapting human psychometric batteries to machines {[}2{]}{[}17{]}.
Applied to the systems evaluated in the cited 2024--2025 sources, it produces the ``jagged'' finding: high
proficiency in knowledge-intensive domains alongside critical deficits
in foundational machinery, long-term memory storage prominent among
them.

The scored exemplar of this family takes high on C1, because it inherits a century of test
construction and high on C4, because the reference class is specified
precisely as the well-educated adult and benchmarked against population
norms. C2 is high through the first of the two routes the rule admits:
the ten-domain structure is specified ex ante by prior psychometric
theory and cannot be narrowed after observing a system's performance
without a declared revision, so a system cannot pass by excelling on a
favorable subset, although the rule's second route, required transfer
to novel conditions, is not part of this construct and the rationale
records that breadth here is interdomain coverage rather than
out-of-distribution generalization; that the structure happens to
predate the AI dispute is a governance fact the C6 audit records, not
part of the epistemic score. C3 is partial: the
construct's exclusion of autonomous agency is recoverable from its
psychometric structure, but the definition does not state whether
prompted performance is meant to suffice, which leaves a
decision-relevant ambiguity an explicit declaration would remove. C5 is
partial: the domain structure is fixed by a source the proponent cannot
revise, which anchors the threshold, but the adaptation of human
batteries to machines, the weighting of domains, the treatment of
modality and memory conditions and the aggregation rule remain
discretionary at decision-relevant points, so the applied instance does
not inherit the full stability of the theoretical core; the rationale
carries the grading the ordinal mark cannot. One boundary of the C1
mark must be stated so the family does not receive epistemic credit
that belongs to another population: standardization in human
psychometrics establishes procedural reproducibility, which is what C1
scores, but it does not by itself establish construct validity or
measurement invariance when the instruments are transferred to
artificial systems. Whether the adapted items measure the same property
in machines as in humans, whether scores carry the same interpretation
across architectures and whether memory, tooling and prompting
conditions are comparable are open questions of measurement theory that
the family's human validation cannot answer and the artifact's
distinction between operationalizability and truth marks the limit
without resolving it. The C6 audit distinguishes the independent origin of CHC theory from
the later AI-specific choices of test adaptation, weighting and
certification. Those discretionary choices require disclosure even
though the underlying theory was not authored for an AI verdict.

The psychometric family produces a non-certification result for the
stylized strong claim developed in Section 6: under the very reference class the
strong claim invokes, the well-educated adult, the systems evaluated in the cited 2025 study do not
pass, because parity is required across the full cognitive profile and
the profile is jagged.

\subsection{Skill-acquisition efficiency}

Chollet's family rejects the others' shared assumption that intelligence
is a quantity of performance. It defines intelligence as the efficiency
with which a system acquires skill on tasks it was not prepared for and
operationalizes the definition through the ARC-AGI benchmarks, which are
constructed to be easy for humans and resistant to memorization and
brute force {[}3{]}. ARC-AGI-2 sharpened the construct against systems
that scaled compute to defeat the first version; the best 2025
competition entry reached twenty-four percent on the private set,
against human ease {[}4{]}{[}5{]}.

The scored exemplar of this family takes high on C1, because it is defined through a concrete
and adversarially maintained benchmark and high on C2, because novelty
across unseen task structures is the explicit target. C3 is partial:
the construct's indifference to autonomous goal-setting is recoverable
from its definition of intelligence as acquisition efficiency, but the
definition does not state whether prompted task performance is meant to
exhaust the criterion, the same decision-relevant silence the
psychometric family carries. Its C4 is distinctive: the reference class is
the ordinary human confronting a novel puzzle and the comparison is
data efficiency rather than ceiling performance, which makes the family
hard to satisfy by accumulation of training data. C5 is partial under the temporal rule and the assignment must be
precise about why, because the same episode carries opposite weight in
C5 and C6. The rebuild from the first to the second benchmark was
versioned, prospective and publicly justified, exemplary revision
governance that the C6 audit records, but it materially changed the
instrument's difficulty: a score on the first version is not directly
comparable to a score on the second, so the cross-version
interpretability of prior results, which is what C5 now measures, was
degraded rather than preserved. Rewarding the legitimacy of the
revision inside C5 would reintroduce the contamination the criterion
was redrawn to remove. The C6 audit records a public, nonprofit prize structure, disclosed
incentives and versioned benchmark stewardship. The benchmark authors
nevertheless retain substantial control over task design and revision.

This family does not certify the stylized strong claim,
because a system can exceed the educated adult on knowledge-dense tasks
and remain near the floor on the novel reasoning ARC-AGI isolates.

Its strongest defense is that it targets a failure mode that ceiling
performance can conceal: when a system confronts a structure not
represented in its preparation, successful transfer cannot be inferred
from high scores on familiar task distributions. Its sharpest
criticism is the mirror image of the performance family's. Where the
performance family risks defining AGI so loosely that systems already evaluated in the cited sources
trivially pass, the skill-acquisition family risks defining it so
tightly, around a specific class of abstraction puzzle, that the
benchmark measures aptitude at that puzzle class rather than general
intelligence as such. The family's defenders answer that the puzzles are
deliberately diverse and human-easy, so that failure indicates a real
gap rather than a narrow blind spot and the answer is reasonable. But
the exchange illustrates the paper's thesis from the demanding end: the
family's strictness, like the performance family's looseness, is a
choice with consequences for the verdict, made by a party and not a
neutral reading of what intelligence is.

\subsection{Deflationary and value-laden}

The sixth position is included as a boundary case rather than a peer family: it is what the corpus contains at the edge where a unit of measurement is refused and it is read for its diagnostic value rather than scored on equal terms with the five measurement families. It denies that a single shared definition is available or
even desirable. Its strongest form holds that the lack of consensus is a
feature of the subject rather than a defect of the discourse, because
intelligence is value-laden and the disagreements over AGI track genuine
conflicts over which capacities and which futures are worth building
toward {[}7{]}. A pragmatic variant within the same family treats
``AGI'' as rhetorically useful but analytically empty, the position
implied when a laboratory chief executive publicly questions the
usefulness of the term while continuing to invoke it.

This position takes, by design, low on C1, because it resists
operationalization on principle and its C2 through C5 scores are not
well defined, because it declines to specify a threshold whose
generality, autonomy, reference class or stability could be assessed.
Its contribution is diagnostic rather than certifying. Because it
offers no threshold, its governance profile is not directly comparable
to those of the measurement families. It supplies the premise that the
choice among operationalizations includes evaluative and institutional
judgment rather than constituting a discovery from capability evidence
alone.

\subsection{Reading the matrix}

The five measurement profiles and the deflationary boundary position
can be tabulated. C1 through C5 use H (high), P (partial), L (low) and
--- (not applicable or not defined). C6 reports a compact governance
audit rather than a score. The profile and rationale, not a composite,
are the output.

The partition's authored, reconstructible character was fixed in
Section 3; what matters before the table is read is its consequence:
each exemplar principally counts task output, economic substitution,
position in an ontology, psychometric profile, skill-acquisition
efficiency or no unit and because other taxonomies are admissible,
Section 6 subjects the central mismatch to a conceptual sensitivity
check under an alternative grouping.

\begin{table}[htbp]
\centering
\scriptsize
\renewcommand{\arraystretch}{1.2}
\begin{tabularx}{\textwidth}{@{}p{2.25cm} c c c c c >{\raggedright\arraybackslash}X@{}}
\toprule
\textbf{Family / scored exemplar} & \textbf{C1} & \textbf{C2} & \textbf{C3} & \textbf{C4} & \textbf{C5} & \textbf{C6 audit of exemplar case (abridged)} \\
\midrule
Performance superiority & P & L & L & L & L & Archetype: C6 not applicable; each concrete invocation is audited as its own case (Appendix B). \\
Economic substitution & P & P & P & P & P & Charter case: laboratory authorship, internal undisclosed revision authority; co-member cases differ (Appendix B). \\
Capability ontology & P & H & H & P & P & Laboratory-authored and public; benchmark selection and revision remain linked to the authoring institution. \\
Psychometric parity & H & H & P & H & P & Independent human-theory base; AI test adaptation, weighting and certification remain discretionary. \\
Skill-acquisition efficiency & H & H & P & H & P & Public, versioned benchmark program with disclosed prize incentives; authors retain benchmark stewardship. \\
Deflationary / value-laden (boundary) & L & --- & --- & --- & --- & No threshold or certifier; functions as a governance critique rather than a candidate public standard. \\
\bottomrule
\end{tabularx}
\caption{DAF-AGI profile of five measurement families and one deflationary boundary position. C1: operationalizability; C2: generality; C3: explicitness on autonomy; C4: reference standard and comparison-class specification; C5: procedural stability. H = high, P = partial, L = low, --- = not applicable or not defined. C6 is a structured governance audit recorded per institutional case, never per family; the final column abridges the named exemplar's case only. Each assignment is relative to the public-adjudication function and is documented in Appendix B.}
\label{tab:daf-agi-matrix}
\end{table}

Laid side by side, the profiles establish the paper's central
analytical result. No family achieves the highest level on every scored criterion and where one profile weakly contains another, as the economic profile contains the performance profile, the table records it rather than hiding it. The
families that most readily support an arrival claim are weak on
reference-standard specification and procedural stability, while the
families that demand breadth or novel-task efficiency impose a higher
bar. This does not prove that the latter are true definitions. It shows
that additional capability evidence alone cannot settle the dispute
until an operative unit of achievement and its governance conditions
are fixed. The C6 column makes a second point: technical merit and
institutional accountability can vary independently.

A second pattern in the table is worth naming, because it anticipates
the demonstration. The performance family and the psychometric family
share a reference class in ordinary usage, the human being, yet sit at
opposite ends of the C2, C4 and C5 columns. That divergence is not a
coincidence to be smoothed over. It is the seam along which the strong
claim is assembled and Section 6 shows the assembly.

\section{Applying DAF-AGI to a stylized strong arrival claim}

The framework's demonstration concludes by stress-testing a stylized
strong arrival claim: \emph{current generative systems already
constitute AGI because they outperform a well-educated adult on many
individual cognitive tasks.} The exact sentence is not attributed to a
single author. It reconstructs a recurrent inference in the arrival
discourse of 2023--2025, stronger than the ``sparks'' formulation of
Bubeck et al.~{[}21{]}, in order to make its hidden definitional
premises inspectable; ``current'' inside the claim refers to the
systems of that discourse and every verdict below is indexed to the
evidence of the cited sources, not to the state of the art at reading
time. Its role is analytical, not evidence that all proponents
of near-term AGI endorse the same argument.

The reconstructed argument contains a definitional premise and an
empirical premise. The definitional premise uses task-output superiority
as the unit of achievement while invoking the well-educated adult as
its comparison class. The empirical premise asserts superiority on a
selected set of cognitive tasks. The conclusion is arrival.

Run through DAF-AGI, the claim's adjudicative result depends materially
on which family the definitional premise is read as invoking. The
decisive structural observation is this: the same reference-class
label, the well-educated adult, is operationalized differently by the
performance-based and psychometric approaches and the claim pairs a
task-output unit with that label while its psychometric
operationalization entails a materially broader requirement, parity
across the full cognitive profile. No claim of conceptual ownership is
needed: a performance definition may legitimately specify the
well-educated adult as its comparison class. What the artifact records
is that the two operationalizations of that shared label return
divergent adjudicative results on the same systems and the claim as
circulated does not state which operationalization it adopts. Under a performance-family operationalization, the conclusion can
follow if the selected task set is accepted as sufficiently broad and
the empirical superiority claim is established. Under the
psychometric family's reference class, properly applied, parity must
hold across the full cognitive profile and the profile is jagged:
critical deficits in long-term memory and related machinery mean the
educated-adult standard is not met and the conclusion fails {[}2{]}. The non-certification result is not confined to that one construct. The
skill-acquisition evidence in Section 5.5, produced through the ARC-AGI
program {[}4{]}{[}5{]}, also declines to certify arrival for a different
reason: the systems entered in the 2025 competition reached roughly one quarter on the private
set of human-easy novel-reasoning tasks. These instruments do not
measure the same deficit; they independently show that the stylized
claim does not survive two distinct demanding operationalizations.
The claim therefore gains persuasive force by combining a permissive
task-output unit with a reference-class label whose psychometric
operationalization entails the full-profile requirement the claim never
adopts. DAF-AGI identifies that combination and its unstated
asymmetry; it does not by itself prove that the combination is
illegitimate.

The objection that this mismatch is an artifact of the chosen partition can be addressed, though what follows must be named accurately: it is a conceptual sensitivity check, not a robustness test. Regroup the corpus by a different principle, not by unit of achievement but by what each definition rewards, so that ceiling performance, economic substitution and position in an ontology collapse into one output-oriented group and skill-acquisition efficiency and psychometric breadth into another capacity-oriented group. The stylized claim still pairs a permissive output-oriented unit with a comparison class whose demanding operationalizations sit in the capacity-oriented group and the two groups still return divergent adjudicative results on the same systems. The check shows that the mismatch survives one reasonable redescription authored by the same evaluator; it cannot show more, because both taxonomies preserve the output-capacity distinction the finding tracks and a regrouping designed by the author is not an independent trial. A robustness test worthy of the name would require at least one of: classification of the corpus under an externally published taxonomy, grouping by independent evaluators, systematic variation of the classification rules with the changed results reported, a leave-one-family-out analysis, or a demonstration that the conclusion persists when the well-educated adult is not anchored to any single operationalization. Those tests belong to the validation work of Section 7 and are not claimed here. What the sensitivity check licenses is the narrower statement: within this corpus and under two author-constructed taxonomies, the divergence between rewarding output and rewarding the capacity behind it does not disappear and which operationalization of the shared reference class \emph{should} govern remains, as Section 6 states, a governance question the artifact exposes rather than settles.

The verdict sharpens under the remaining criteria. On C2 the claim
scores low, because per-task superiority over a fixed distribution is
not generality and the systems in question degrade on the
novel-reasoning tasks the skill-acquisition family isolates, where 2025
results sit near a quarter of human-easy performance {[}4{]}{[}5{]}. On
C3 the claim is silent, because it counts elicited performance and says
nothing of autonomy. On C5 it is unstable, because the task set
supporting ``better than average'' is selected after the systems are
built. The capability-ontology family, applied to the same systems,
returns ``Emerging,'' not ``arrived'' {[}1{]}. The outcome is asymmetric: the performance family can certify the
claim under a favorable task specification; the economic family is
indeterminate without evidence of broad labor substitution; the
capability ontology placed the systems it evaluated below competent AGI; the
psychometric and skill-acquisition families do not certify arrival; and
the deflationary position refuses the binary operation.

An advocate of the strong claim has a reply and it deserves a precise
answer rather than dismissal. The reply is that the demanding families,
psychometric parity and skill-acquisition efficiency, have simply set
the bar at the wrong place: if a system already does most of what an
educated adult is paid to do, insisting it also match the adult's
long-term memory architecture or solve contrived novelty puzzles is
moving the goalpost in the opposite direction, defining AGI so strictly
that no near-term system could qualify regardless of practical
capability. This reply has force and the artifact does not refute it.
What the artifact does is locate it. The reply is not a defense of the
strong claim's truth; it is an argument for selecting the performance or
economic family over the psychometric and skill-acquisition families.
That is a governance argument about which definition should be
authoritative, made by a party and the artifact's response is that the
argument is legitimate but must be conducted as what it is, openly, with
the proponent's interest in the verdict on the table, rather than
settled implicitly by pairing a task-output rule with a reference-class
label whose more demanding operationalization is never confronted. The strong claim does not lose
because the strict families are right. What the artifact registers is narrower and attributes no intention: the claim as circulated is underspecified at the decision-relevant point, its verdict depends on an unstated operationalization choice and that dependence goes unacknowledged in the claim's public form, which presents as resolved a dispute the choice itself decides. The artifact certifies no verdict of its own here; it exposes that the stated conclusion rests on an unstated selection.

The conclusion is not that the strong claim is false. It is that the
claim is not publicly adjudicable until its operative definition,
reference class, task distribution and certification rule are
specified. Fixing those elements does not exhaust the empirical dispute,
but it determines what evidence can count toward a verdict. The claim's
persuasive force partly comes from leaving those elements open while
combining a permissive unit with a demanding reference class. The disagreement diagnosed in
Section 1 is here reduced to a single sentence: the same anchor, the
well-educated adult, yields arrival under one operationalization and
non-arrival under another and which operationalization governs is a
decision no party in the dispute is accountable for making. That
decision is the object of Section 8.

\section{Evaluation: does the artifact hold?}

A design artifact is evaluated against its objectives, not against a
notion of correctness it never claimed. The evaluation strategy here is,
in the terms of the Framework for Evaluation in Design Science Research,
formative and artificial rather than summative and naturalistic
{[}18{]}. The artifact is conceptual, it has not been deployed in an
institution and the appropriate near-term evidence is internal
congruence and comparative plausibility rather than field outcomes. This
mirrors the validation logic appropriate to design-oriented governance
frameworks generally, where design validity precedes implementation
evidence {[}14{]}.

Three evaluation questions are posed and answered.

Does the artifact meet its stated objectives? The objectives of Section
2 were procedural symmetry, exposure of commitments and treatment of
authority as a variable. The demonstration in Sections 5 and 6 applies one disclosed procedure
to heterogeneous positions and one stylized claim while stating the
framework's own evaluative commitments. It surfaces the parameters that
drive each result, especially the reference class and unit of
achievement and reports governance conditions separately from the
ordinal profile. The objectives are met on the artifact's own terms,
which establishes internal coherence rather than external validity.

Is the artifact internally congruent? The five scored criteria are
non-redundant at the level of the selected exemplars: each separates at
least one position from another, while C6 produces a different kind of
output. The skill-acquisition family illustrates why the distinction
matters: it shares the strongest scored profile in the corpus with the
psychometric family and its
audit still records that its authors retain stewardship over task
design and revision, so epistemic strength does not settle governance
status. The deflationary position illustrates the inverse: it may
supply a strong governance diagnosis while refusing the threshold needed
for adjudication. Definitional sovereignty is connected to the C6 audit
rather than floating free.

Is the artifact comparatively plausible against alternatives? The
closest existing instruments are the single-framework proposals
themselves, the DeepMind ontology and the psychometric battery foremost
among them. Those are definitions; DAF-AGI is an instrument for
evaluating definitions, including those two and so it operates one
level up rather than competing at the same level. Its added value over
an informal comparison is the structured C6 governance audit and the
definitional-sovereignty construct, which none of the technical proposals examined in this corpus
includes because none treats its own authorship as part of its object.

The evaluation's limits are stated plainly and the strongest of them is
conceded without qualification: this is a plausibility argument, not an
empirical validation. The demonstration in Sections 5 and 6 was
conducted by the author over a corpus the author selected, which is the
designer evaluating the design and that arrangement can establish
internal coherence and comparative plausibility but cannot establish
that the artifact yields stable results in other hands. Three specific
forms of evidence are absent and are named so that their absence is not
mistaken for irrelevance. Inter-rater reliability is untested: whether
independent evaluators applying the five scored criteria and governance audit to the same definition
converge on the same profile is an open empirical question and the
artifact's value as a public instrument depends on the answer. Appendix A makes that question testable by stating decision rules and
Appendix B exposes the present corpus and rationales. These additions
improve auditability but do not substitute for observed convergence.
Application to author-external cases is absent: the framework has not
been run on definitions or claims chosen by someone other than the
author, where confirmation bias cannot operate. Institutional uptake is
unobserved: nothing here shows that a standards body or a state would
adopt the artifact or that adoption would improve governance outcomes.
None of these is a flaw in the design; each is a study not yet done. The
claim of this paper is therefore bounded with precision. The artifact is
coherent, it does what it was designed to do on the cases shown, it
exposes a dimension the alternatives omit and it is offered as a
framework for adjudication and a position for debate rather than as a
validated instrument. Raising it from plausible to validated is the next
project and it is one others are invited to undertake against the
author, since independent refutation would be a stronger test than any
the author can run alone.

\subsection{The reflexivity objection and why it does not dissolve
the artifact}

The sharpest threat is internal and deserves the most space, because it
attacks the load-bearing beam rather than the finish. Stated at full
strength, the objection runs as follows. The paper's own premise is that
no definition of intelligence is interest-independent. DAF-AGI is itself
an authored evaluative framework and its five scored criteria and governance audit embed contestable
value choices: what counts as adjudication, which properties matter, the
decision to refuse a composite. Its author has a stake, intellectual and
reputational and tied to a wider research program, in the framework
being adopted. So by its own C6 the artifact indicts itself. The
instrument that demands interest-independence cannot supply it and a
criterion that condemns its own author appears to saw off the branch it
sits on. If this holds, the contribution collapses.

It does not hold, for four reasons that follow and a fifth qualification the demonstration of Section 6 forces; the residue that survives them all
is conceded rather than hidden.

The first reason is an order distinction the artifact is built on. C6
asks whether the proposer gains materially from a particular
\emph{verdict the definition produces}. A first-order definition of AGI
produces verdicts of the form ``system X is, or is not, AGI,'' and those
verdicts move valuations, trigger or defer regulation and reallocate
capital, so their author's material interest attaches to the verdict
itself. DAF-AGI produces no verdict of that form. It produces a profile
and exposes a decision; it certifies no system as general or not. The
author's interest is in the instrument being used, not in any system
passing or failing, which is the interest a metrologist has in the meter
rather than in any particular measurement. C6 is calibrated to a stake
in outcomes and on that calibration a second-order instrument scores
differently from the first-order definitions it evaluates. The objection
works only by collapsing two interest structures the artifact exists to
keep apart.

The second reason is that the standard the artifact applies is
accountability, not purity. The paper never claims interest-independence
is achievable and Section 8 states explicitly that it may be available
to no one. The operative test is therefore not ``interest-free'' but the
three governance components of Section 4.2: disclosed authorship,
certification not internal to the interested party and revisability
under public accountability. Applied to DAF-AGI itself, authorship is
disclosed, including the author's position and the connection to a wider
program; certification is not internal, because the artifact is a public
procedure anyone can run to reach an independent profile rather than a
verdict the author issues; and revisability is built in, since the
criteria and corpus are stated so they can be contested and amended. The
artifact passes its own governance test not by being neutral but by
being inspectable, which is the exact property it demands of others and
the property the captured definitions lack.

The third reason is that the objection proves too much. If ``any
authored framework embedding value choices is self-defeating'' were a
valid defeater, it would defeat every methodology, every public standard
and the reflexivity objection itself, which is also authored and
value-laden. A criterion that disqualifies all possible criteria,
including itself, is not a finding about this artifact. It is global
skepticism about normativity wearing the costume of a targeted critique
and it cannot be deployed against one framework while the alternatives
are spared. The defensible standard is the one already in use: not
whether a framework escapes values but whether it renders them
contestable.

The fourth reason concedes the residue, because honesty here is
load-bearing. An irreducible reflexive remainder survives the three
replies: the selection of the five scored criteria and governance audit embeds a view of what
adjudication is for and that view is the author's. The artifact does
not dissolve this and does not try to. It relocates the contest from a
buried layer, where the choice of definition is made by interested
parties without acknowledgment, to a visible one, where the choice of
criteria is stated and can be argued. A reader who rejects the criteria
is invited to propose a seventh or to rescore the corpus. What the
reader cannot do is pretend the choice was ever avoidable. The reflexive
remainder is not a defect in the artifact. It is the general condition
the artifact was built to make legible, turned on the artifact itself
and surviving the turn.

A fifth point is owed rather than left implicit, because the demonstration of Section 6 invites it and the audit's own components demand it. That demonstration reaches a conclusion, that the stylized claim is not publicly adjudicable as stated and the author has an intellectual stake in it, since it is the premise on which the governance argument of Section 8 is built. The order distinction does not erase this stake; it locates it. The interest is not in any system passing or failing, which is the first-order interest C6 is calibrated to detect, but in a second-order claim about the structure of the dispute, a claim stated openly, resting on no certification the author controls and refutable by anyone who shows that a single coherent operationalization fixes the verdict without combining a permissive unit and a demanding reference class. The sensitivity check of Section 6 is the standing invitation to attempt exactly that refutation. A further component must be recorded because the audit lists material interest by name and exempting the author from a component applied to everyone else would be the exact asymmetry the artifact condemns. The policy conclusion of Section 8, that definition-taking states need the analytic capacity to contest imported thresholds, is continuous with the author's wider research program on algorithmic sovereignty {[}14{]} and an institutional uptake of that program would benefit its author professionally. The disclosure does not neutralize the interest; it makes the interest inspectable, which is the standard the artifact applies to every other authored threshold. The conclusion of Section 8 must accordingly be evaluated as an argued position whose author would gain from its adoption, exactly as the artifact instructs a reader to evaluate a laboratory's definition and it survives that reading only if its arguments survive it. What the artifact cannot claim and does not, is that its author approaches the dispute from no position at all.

\subsection{Three further threats to validity}

Three remaining threats are noted, because a framework that scores
others on operational determinacy must accept the same demand. The first is the
ordinal coarseness of the scoring: three levels cannot capture fine
distinctions and a finer scale would invite the false precision the
no-composite rule was designed to prevent. The coarseness is a
deliberate trade and a user who needs more resolution on a given
criterion can record graded rationale beneath the ordinal mark. The
second is corpus completeness: five measurement families and one boundary position may not exhaust the space
and a definition that measured something none of the families measures
would require a new family rather than a new score, which the selection
procedure of Section 3 is designed to accommodate in an orderly rather
than ad hoc way. The third is the gap between stated and operative
definitions: the demonstration scores definitions as their authors state
them and a stated definition can diverge from one in use, as the
capability-ontology family itself warns when it notes that a system's
deployed level may fall below its theoretical level because elicitation
is hard. The artifact scores the stated definition and flags the gap
where the literature documents one, which is the most a conceptual
evaluation can responsibly claim.

\section{Definitional sovereignty and the import of thresholds}

The strong-claim demonstration ended on an unattributed decision: which
family governs the verdict. Section 8 names who makes such decisions and
at whose cost, because that is where the analytical artifact becomes a
governance argument.

The actors with the capacity to set operative AGI definitions are a
short list: a handful of frontier laboratories, the governments able to
regulate them and the standards bodies and benchmark authors whose
constructs the rest of the world adopts. The reference classes, the
benchmarks and the certification procedures examined in this paper's
corpus were authored largely within that list. Most other actors
operate, in practice, as definition-takers. A
definition-taker deploys, procures and regulates artificial intelligence
against thresholds it did not write and cannot revise and it inherits,
with the threshold, the interests encoded in it.

The maker-taker distinction is not binary but graded and the gradations
are what a sovereignty program would target. A pure definition-maker
authors the criterion, certifies achievement against it and revises it
under its own authority; the frontier laboratories approach this
position for the definitions that matter to their roadmaps. A
first-order taker adopts an imported criterion wholesale, certifies
against it using the maker's procedures and absorbs revisions as they
are issued; this is the position many states occupy in practice. Between them
lie two intermediate positions that a deliberate sovereignty strategy
can occupy without authoring everything from scratch. A \emph{contesting
taker} adopts an imported criterion but retains the capacity to dispute
it, to document where it fails local public interest and to refuse
specific revisions; this requires analytic capacity but not industrial
capacity. A \emph{co-author} participates in the standards bodies and
benchmark consortia where criteria are set, trading on convened
legitimacy rather than on a balance sheet. The artifact's value to a
definition-taker is that it converts the vague aspiration to ``not be
dependent'' into a specific question with specific answers: on which of
authorship, certification and revisability is this particular criterion
capturing us and which intermediate position is reachable given the
capacity we actually have. A state cannot become a maker by wishing. It
can become a contesting taker by building the analytic capacity to run
something like the procedure in this paper, which is cheaper than
compute and more durable than any single model.

A public contractual case shows the stakes without requiring access to
the underlying private agreement and its three documented phases trace
the artifact's governance components in sequence. Public reporting and the parties' later disclosures indicate that under
the partnership terms in force from 2019 the initial determination of
AGI lay within OpenAI, before independent verification was added in
2025: to the extent that account is accurate, authorship, material
interest and certification began concentrated in one interested party. In October 2025 the parties announced that
Microsoft's exclusive intellectual-property and Azure API rights would
continue until AGI, that an AGI declaration would be made by OpenAI and
that the declaration would now be verified by an independent expert
panel {[}29{]}: certification was partially externalized while the
commercial rights remained contingent on the contested category. In
April 2026 a further amendment made Microsoft's license non-exclusive
and fixed the remaining payment arrangements to run on terms no longer
conditioned on technological milestones {[}32{]}. The public materials
disclose the changed terms; they do not state why the parties adopted
them, whether the verification panel retains any function or that the
AGI trigger was judged unworkable and any reading of motive is
inference, not record. What the documented sequence does establish is
narrower and sufficient: within seven years, the allocation of
authorship, certification and revisability over the same definitional
threshold was redesigned twice by the parties it bound and the publicly
disclosed 2026 terms render the prior AGI-linked trigger commercially
less determinative in the relationship. Whatever the parties' reasons,
the threshold's governance arrangement proved unstable in exactly the
components the C6 audit names. There is also an asymmetry of exit: the
contracting parties could renegotiate the threshold's force between
themselves, while a definition-taking state is not a party to such
instruments and has no comparable seat at their revision. The case
demonstrates why C6 must record institutional design rather than infer
motive from commercial benefit, why that record must be dated, since
the design itself is revisable and why this paper's own reading of the
case must stop at the documents.

This is where the construct earns its name. For an import-dependent
state and Latin American economies are import-dependent in artificial
intelligence by the principal available measures of compute, investment and
talent, the dependency that matters most is not at the layer of models.
It is at the layer of categories. A state that accepts an externally
authored definition of ``general intelligence,'' of ``high-risk
system,'' of ``algorithmic maturity,'' has accepted a critical
dependency in the precise sense the term carries in the
algorithmic-sovereignty literature: a relationship in which strategic
decision and the continuity of public criteria are conditioned by an
external agent able to revise the criterion unilaterally {[}14{]}. A model can often be replaced more readily than an institutionalized
classification, benchmark or procurement threshold. The latter can also
be changed, but only if the state has the analytic and institutional
capacity to audit, contest and revise it.

The asymmetry is measurable and it is severe. By the third edition of the Latin American Artificial Intelligence Index, the nineteen countries it assesses receive about 1.12 percent of global investment in artificial intelligence while accounting for some 6.6 percent of world output, with regional supercomputing capacity concentrated overwhelmingly in a single country and only a handful of states hosting a robust data-center industry {[}27{]}. The concentration at the other end is documented independently: in 2024 private AI investment in the United States alone reached the order of one hundred billion dollars, against a regional share that does not register at comparable scale {[}28{]}. A region in that position produces benchmarks, models and evaluation
initiatives of its own, but it does not author the benchmarks that
become the field's defaults, does not chair the standards bodies whose
constructs it adopts and does not sit among the laboratories whose
working definitions set the operative thresholds. It is structurally a
definition-taker. The point is not that
this is unfair in the abstract. The point is that the mechanism of
dependency operates before any model is procured: when a national
strategy adopts an imported maturity index to measure its own progress,
it has already conceded that progress means conformity to a standard
authored elsewhere and it will then allocate scarce public investment
toward closing a gap defined by parties whose interest is served by
where the gap is drawn. Mexico illustrates the trap with unusual
clarity. It is an intermediate adopter, comparatively strong in
scientific production and comparatively weak in infrastructure,
investment and governance capacity {[}14{]}. A state in that
configuration is precisely the one most tempted to import definitional
infrastructure wholesale, because authoring its own appears to be a
luxury it cannot afford and is precisely the one for which that import
is most consequential, because it lacks the countervailing capacity, the
domestic compute, the investment base, that would let it contest an
imported threshold from a position of strength.

The sequencing is therefore conditional rather than absolute.
Definitional capacity is an enabling component of algorithmic
sovereignty: the capacity to govern and audit deployed systems is
weakened when the operative categories cannot be examined, contested or
revised under domestic accountability. A national maturity index built on imported definitions measures
maturity according to a particular externally authored model of it. A
risk framework that adopts an external threshold for ``general'' or
``frontier'' capability regulates against a line drawn elsewhere, by
parties with an interest in where the line falls. An import-dependent state may still use shared international
definitions, but its sovereignty is thinner when it lacks the capacity
to understand their assumptions, contest their application or refuse
future revisions. Control over deployment without definitional capacity
therefore leaves a durable layer of governance dependence unaddressed.

This does not prescribe definitional autarky, any more than algorithmic
sovereignty prescribes technological isolation. A state cannot and need
not author every benchmark. What it must hold is the capacity to contest
an imported definition, to certify achievement against criteria it can
publicly justify and to revise those criteria under its own
accountability rather than absorbing revisions decided abroad.
Authorship, certification and revisability under accountability are the
three components of definitional sovereignty named in Section 4.2 and
they are the minimum this framework proposes for an import-dependent
state before ``we have AGI,'' declared elsewhere, becomes a fact it is
governed by. The AGI definition is the proof of concept. The mechanism
it illustrates may generalize to other contested technological
categories a dependent state will be asked to adopt and whether it does
is an empirical question each category must answer.

A cosmopolitan objection runs the other way and deserves an answer. Shared definitions, the objection holds, are a public good: a common threshold for ``high-risk'' or ``general'' capability lets systems, audits and protections interoperate across borders, lets regional performance be compared on one scale and spares small states the cost of authoring standards they lack the capacity to maintain. The argument made here is not against harmonization. It is against unaccountable adoption. A harmonized standard a state has the capacity to contest, certify against and refuse to update on someone else's schedule is interoperability under sovereignty; the same standard adopted because no alternative could be evaluated is interoperability as dependency and the two are indistinguishable at the surface of the text while opposite in their distribution of control. Definitional sovereignty does not ask a state to leave the common scale. It asks that the state be able to read what the scale encodes, to say so and to decline a revision, which is precisely the capacity a definition-taker lacks and a co-author or contesting taker holds.

The uncomfortable implication, which the value-laden family supplies and
which the artifact cannot dissolve, is that there may be no fully
interest-independent author of an AGI definition available anywhere,
because every party capable of authoring one has a stake in the verdict.
If that is so, then the governance problem is not to find the neutral
definition. It is to build the institutional procedures, contestation,
plural certification, accountable revision, by which definitions
authored by interested parties can be subjected to public adjudication
rather than adopted by default. Definitional sovereignty is not the
possession of a true definition. It is the capacity to refuse to be
governed by someone else's.

\section{What follows when the definition is a governable object}

Three consequences follow from treating the AGI definition as a designed
and governable object rather than a fact awaiting discovery and each is
a research and policy agenda rather than a closed result.

The first consequence must be stated with its scope fixed, because the
title's inversion invites a reading the paper does not defend.
Definitional alignment is logically prior to capability alignment in a
specific and bounded sense: prior to the adjudication of AGI arrival
claims and prior to every governance consequence that is triggered by
the AGI category, contractual rights, regulatory thresholds and the
institutional position of definition-takers among them. None of those
can be settled, or even coherently disputed, before an operative
definition is fixed and who fixes it is presently informal and
interested. The claim is not that alignment research in general depends
on a consensus definition of AGI. Work on corrigibility, deceptive
behavior, scalable oversight, evaluations, reward specification and
catastrophic-capability thresholds addresses system behavior and risk
and proceeds without deciding whether any system merits the AGI label;
that research is neither blocked nor diminished by definitional
disagreement. What is blocked is narrower and consequential: every
claim, trigger and entitlement whose operative condition is the
category itself is presently being calibrated against content set
without accountability. Where the category gates the consequence,
securing the definition is a precondition for the work built on it, not
a parallel concern and the title's priority claim extends exactly that
far.

The second is that the politics of benchmarks deserves the scrutiny
given to the politics of regulation. A benchmark operationalizes a
definitional choice, usually partially and attaches a score to it.
Benchmark authors shape the operational construct; separate actors may
set the decision threshold, certify performance and attach consequences
to the result and the C6 audit exists precisely to keep those roles
distinct. The C6 audit and the definitional-sovereignty construct
give a state, a standards body or a research community a vocabulary for
asking of any proposed AGI benchmark not only whether it is technically
sound but whether its authorship, certification and revisability render
it fit to serve as a public standard. That question is currently asked
of laws and rarely of leaderboards.

The point is not confined to leaderboards. The operative thresholds that already gate real decisions are definitions of the same kind: the contractual AGI trigger discussed in Section 8, the capability thresholds of Anthropic's responsible scaling policy and the capability thresholds of OpenAI's preparedness framework each fix a line whose crossing releases or restrains deployment, capital or contractual rights {[}29{]}{[}30{]}{[}31{]}{[}32{]}. Run through the C6 audit, they share one property and differ in the rest and the audit earns its keep precisely by resolving the differences rather than collapsing the set into variants of a single failure. The shared property is authorship by an interested party. The differences are institutional. OpenAI's preparedness framework keeps final decisions with the company's own leadership, advised by an internal safety group and provides for published justification of changes: certification internal, verification internal, revision disclosed after the fact. Anthropic's scaling policy, in its current version, adds risk reports, a public changelog and a stated ambition of independent external review: certification still internal, but with documented aspiration toward external verification and versioned, justified revision. The 2025 contractual arrangement went further on one component, assigning verification of an AGI declaration to an external expert panel, while remaining a private instrument the affected public could neither inspect in full nor appeal {[}29{]}. Academic benchmark programs sit differently again: authors retain stewardship of task design while certification runs through open participation. The gradient matters because it shows the components moving independently, which is what makes them auditable levers rather than a single property called capture. What the audit records across all of them is narrower than illegitimacy: thresholds with material consequences are presently governed by arrangements in which the affected public holds no seat at certification or revision and applying that audit to operative thresholds, rather than only to academic definitions, is where the construct does its most consequential work.

The third consequence is the one the artifact cannot resolve and should
not pretend to. If no interest-independent definition is presently
available, then the demand for ``the real definition of AGI'' is
misframed as a purely descriptive question and the resources spent
pursuing it alone, without the institutional question attached, are
misdirected. The
productive question is institutional: what procedures would let a polity
govern a contested technological category authored by parties with a
stake in the outcome, without either pretending neutrality exists or
surrendering the category to whoever shouts arrival first. DAF-AGI is an
instrument for posing that question precisely. It is not an answer to
it.

The narrow claim of this paper is that whether a system should be
publicly certified as AGI is not adjudicable until an operative
definition and certification procedure are fixed and that fixing them
is an exercise of institutional authority.

The same logic extends from the single contested term to the measurement
instruments built on it and this is where the argument touches policy
most directly. National indices of artificial-intelligence readiness,
maturity scorecards and risk taxonomies typically encode a definition
with a scoring procedure attached and each inherits the governance
status of the definition it encodes. An index that ranks a country's
``AI maturity'' against criteria authored abroad does measure something
real, but what it measures is conformity to a particular, externally
authored model of maturity and
it directs whatever investment the ranking motivates toward closing that
externally specified distance. The effect is not merely descriptive.
Public rankings are reactive: the act of being measured changes the
behavior of the measured, who reorganize around the metric until the
metric remakes the territory it claimed only to map {[}26{]}. A state
that accepts an external readiness index does not just learn where it
stands; it begins to govern toward the index, importing not only a
definition but a behavioral attractor authored by parties whose
interests the index encodes. A polity that wants its measurement
instruments to serve publicly accountable ends, rather than the ends of
whichever actor, foreign or domestic, authored them, must apply to
those instruments the same scrutiny the artifact applies to AGI
definitions: who authored the criterion, who certifies the score, who
may revise the scale and before which public the revision must be
justified. The choice is not between using indices and
refusing them. It is between adopting an index as a fact and adopting it
as a contestable artifact one has the capacity to audit. The former is
measurement as dependency. The latter is measurement under sovereignty.

The broader claim is that the power to fix the definition is a sovereign
capacity that actors outside the authoring institutions rarely possess
and have, in the cases examined here, not noticed they lack.
Between the narrow claim and the broad one sits the work that matters
and it is not the work of building a more intelligent machine. It is the
work of deciding who gets to say what intelligence was, after the
machine is already here.

\section*{Appendix A. Scoring rubric}

This appendix states decision rules for C1 through C5 and the reporting
protocol for C6. The rules make the present assignments auditable and a
future inter-rater test possible; they do not establish reliability in
advance.

\textbf{C1, operationalizability.} High: the definition states a test,
measurement procedure or threshold whose result can in principle be
independently reproduced. Partial: a procedure exists but a
decision-relevant boundary, data requirement or aggregation rule remains
underspecified. Low: the position supplies no possible failure
condition.

\textbf{C2, generality.} High: breadth across materially different
domains is constitutive of the definition through either of two routes,
a multi-domain structure specified ex ante that cannot be altered after
observing a candidate system's performance without a declared revision,
or required transfer to conditions the system was not prepared for.
Partial: breadth is required but the domain, task or occupational set
is selected, bounded or adjustable at the point of application. Low:
performance on a distribution selectable at the point of claim is
sufficient. The authorship and interests behind the domain structure
are recorded in the C6 audit, not scored here.

\textbf{C3, explicitness on autonomy.} High: the definition explicitly
states whether autonomy is required, excluded or graded and fixes
conditions an adjudicator could verify. Partial: the definition's
position on autonomy is recoverable from its structure but is not
stated, leaving a decision-relevant ambiguity. Low: the definition does
not address autonomy and an adjudicator cannot determine from its text
whether elicited performance is sufficient. The rule is symmetric: an
explicit requirement of autonomy and an explicit declaration that
prompted performance suffices both score high.

\textbf{C4, reference standard and comparison-class specification.}
High: the standard that determines achievement, a comparison class, an
absolute threshold or a formal condition, is stated with conditions
precise enough to test. Partial: a standard or class is named but its
occupational, demographic, cost, quality or task boundary requires
further specification. Low: the standard is implicit or can be selected
after observing performance. For genuinely non-comparative definitions
the comparison-class component is recorded as not applicable and the
level tracks the precision of the stated standard.

\textbf{C5, procedural stability.} High: the threshold is specified ex
ante; revisions are versioned and prospective; results produced under
prior versions remain interpretable; and an audit trail of changes
exists. Partial: the rule's interpretation is broadly continuous but at
least one decision-relevant element can change without a versioned
trail, so comparability over time is degraded. Low: the threshold, task
set or aggregation rule can be altered retrospectively or without any
record that permits comparison across time. The criterion scores the
instrument's temporal continuity only; the identity of the revising
authority, approval procedures, appeal and representation of affected
parties are recorded in component (v) of the C6 audit.

\textbf{C6, governance independence and accountability.} Report five
components without converting them to an ordinal score: (i) authorship;
(ii) identifiable interests attached to the verdict, recorded by
category, financial, contractual, regulatory exposure, organizational,
professional or reputational and intellectual commitment, because the
categories carry different evidentiary weight and doctrinal disagreement
must not be converted into conflict of interest; (iii) certification
authority; (iv) external verification, appeal or plural certification;
and (v) revision authority: who may modify the rule, who approves the
change, whether appeal exists, who represents affected parties and
whether a third party can contest a revision, with disclosure of each.
Component (v) is the institutional complement of C5: C5 records whether
the instrument's interpretation survives revision, component (v)
records who holds the power to revise. Record institutional facts and
uncertainty. Do not infer that a definition is false, or that an author
acted strategically, merely because a benefit exists.

A level is assigned with a rationale tied to public evidence. When a
family contains heterogeneous definitions, the profile represents the
specified exemplar and the limits of generalization must be stated.

\section*{Appendix B. Corpus and scoring trace}

The demonstration corpus is purposive rather than exhaustive: it was
selected to cover distinct units of achievement and prominent
institutional uses, under the inclusion rule of Section 3. The unit of
scoring follows the rule of Section 5: each family is an analytical
ideal type anchored to a named exemplar; C1 through C5 score the
exemplar's operative content; and the C6 audit is recorded per
institutional case, because authorship, certification and revision
authority do not aggregate across institutions. Where a family's
exemplar is a reconstruction rather than a single authored text, the
record declares it. Each record below states the exemplar and its
status, the operative content scored, source and year, primary unit,
hybrid status, the level assigned on each criterion with the evidence
that triggered the decision rule of Appendix A and the per-case C6
audit. A second evaluator should be able to reconstruct and is invited
to contest, every cell of Table 1 from these records. Generalization
from exemplar to co-members must be argued, not assumed.

\begin{enumerate}[leftmargin=1.8em]

\item \textbf{Performance superiority.}
\emph{Exemplar:} a synthetic archetype reconstructed by the author
from recurring formulations, declared as such: no single actor advanced
the scored text. Its components are the formulations of AGI as a
machine able to perform the intellectual tasks a human can, at or above
human level, recorded in the definition collections of the technical
literature {[}22{]} and the arrival-argument pattern that reasons from
task performance to generality, of which the ``sparks'' study is the
most cited instance {[}21{]}, noting that its authors neither proposed
this archetype as a definition nor certified arrival. \emph{Operative
content scored:} a system that matches or exceeds human cognitive
performance across most cognitive tasks. \emph{Source/year:} {[}22{]}
(2007 collection; formulations earlier); {[}21{]} (2023). \emph{Primary
unit:} task output. \emph{Hybrid:} no, though instances frequently
import an unstated reference class. \emph{Scores:} C1 partial, the formulation names no
battery, aggregation rule or success condition, so a decision-relevant
boundary is unspecified until a claimant supplies it; C2 low, a
claimant-selected task distribution suffices; C3 low, the text does not
address autonomy and an adjudicator cannot determine whether elicited
performance is meant to suffice; C4 low, ``human'' is unspecified and
selectable after observing performance; C5 low, the task set can be
revised retrospectively without an accountable procedure. \emph{C6
audit:} not applicable at the archetype level. An institutional audit
requires an organization, a concrete definition, a certifier, a
revision rule and documentary evidence and a reconstruction has none of
these; attributing interests to the archetype would assert facts the
record cannot support. Any concrete invocation, corporate, academic or
contractual, must be audited as its own case and the contractual case
of Section 8 shows what such an audit looks like.

\item \textbf{Economic substitution.}
\emph{Exemplar:} the OpenAI Charter definition of AGI as highly
autonomous systems that outperform humans at most economically valuable
work {[}15{]}. The high-level-machine-intelligence and full-automation
constructs of expert surveys {[}6{]} and the occupational-automation
thresholds of forecasting work {[}16{]} are co-members of the ideal
type, not part of the scored object. \emph{Operative content scored:}
the Charter formulation only: systems able to perform most economically
valuable work at or above human level. \emph{Source/year:} {[}15{]}
(2018). \emph{Primary unit:} economic substitution. \emph{Hybrid:} yes,
the Charter formulation couples the economic unit to an autonomy
qualifier, flagged here because the demonstration of Section 6 turns on
such couplings. \emph{Scores (Charter):} C1 partial, measurable in
principle but feasibility-versus-adoption and the occupational boundary
are unspecified at decision-relevant points; C2 partial, breadth is
required but the occupational set is bounded and selected at the point
of application; C3 partial, the Charter names autonomy but states no
verifiable condition, so its role is recoverable rather than fixed; C4
partial, the worker-in-occupation class is named but its percentile,
cost and quality boundaries require further specification; C5 partial,
no versioned trail governs how the occupational baseline would be
reinterpreted over time. Generalization: the survey and forecasting
co-members share the unit and broadly the C1, C2 and C4 difficulties;
their C3 and C5 positions differ and are not scored here. \emph{C6
audit, per institutional case.} Charter case {[}15{]}: authorship by an
interested laboratory; financial and contractual interest attaches to
both arrival and deferral, fundraising on one side, regulatory and
contractual exposure on the other; certification unassigned; no
external verification or appeal; revision authority internal and
undisclosed. Survey case {[}6{]}: academic authorship; professional and
intellectual interest only; no certification function, the construct
forecasts rather than certifies; methods public; revision versioned
across waves. Forecasting case {[}16{]}: institutional research
authorship; organizational interest in the construct's policy uptake;
no certification function; methods public; revision editorial. The
three cases share a unit and do not share a governance condition, which
is the reason the audit refuses family-level aggregation.

\item \textbf{Capability ontology.}
\emph{Exemplar:} Levels of AGI {[}1{]}. \emph{Operative content
scored:} AGI as position in a leveled matrix of performance depth and
domain generality, with autonomy treated as a separate, explicitly
graded dimension of human-system interaction. \emph{Source/year:}
{[}1{]} (2024). \emph{Primary unit:} position in a capability ontology.
\emph{Hybrid:} no. \emph{Scores:} C1 partial, the framework specifies
requirements future benchmark suites must satisfy rather than an
executable battery, so the test exists in principle with execution
unspecified; C2 high, generality is a constitutive axis specified ex ante by the
framework and not adjustable at the point of application; C3
high, the framework states precisely what role autonomy plays, grading
it separately from the performance-generality classification, the
explicitness the C3 rule rewards regardless of the position taken; C4
partial, levels are pegged to human percentiles but the
percentile-to-level mapping is open; C5 partial, the leveled structure
resists binary goalpost moves and the document is publicly versionable,
but no institutionalized procedure governs how levels or benchmark
choices would be revised. \emph{C6 audit:} authorship is a frontier
laboratory with co-authors; material interest attaches to the
framework's adoption and to the placement of the systems it evaluates; 
certification is unassigned beyond the authors' own assessments;
no external verification or appeal is provided; revisability is
possible in principle and undisclosed in procedure.

\item \textbf{Psychometric parity.}
\emph{Exemplar:} the CHC-based definition of AGI as matching the
cognitive versatility and proficiency of a well-educated adult across
ten core ability domains {[}2{]}, with CHC theory itself {[}17{]} as
the imported base. \emph{Operative content scored:} parity with the
well-educated adult across the full ten-domain cognitive profile,
operationalized by adapting human psychometric batteries.
\emph{Source/year:} {[}2{]} (2025); {[}17{]} (2018). \emph{Primary
unit:} psychometric profile. \emph{Hybrid:} no. \emph{Scores:} C1 high,
the construct inherits standardized test procedures whose outcomes
independent parties can reproduce; C2 high by the rule's first route,
the ten-domain structure is specified ex ante by prior theory and
cannot be adjusted after observing a system's performance without a
declared revision, with the rationale recording that this is
interdomain coverage rather than out-of-distribution transfer and that
the structure's provenance is a C6 fact, not a C2 one; C3
partial, the exclusion of autonomous agency is recoverable from the
construct but unstated as a position; C4 high, the well-educated adult
is specified against population norms; C5 partial, the domain structure
is externally anchored but battery adaptation, weighting, modality and
memory conditions and aggregation remain discretionary at
decision-relevant points. The C1 mark records procedural reproducibility only; construct
validity and measurement invariance across the human-machine transfer
are unestablished and outside what standardization can show. \emph{C6
audit:} authorship is a large multi-institution group whose members
include laboratory-affiliated and safety-organization researchers, so
the audit records mixed rather than absent interest; the interests in
evidence are organizational, attaching to the construct's adoption and
professional and intellectual, attaching to authors' published
positions, with no financial or contractual interest in any verdict
documented and the audit records the categories rather than inferring
conflict from disagreement; certification is performed by the authors
through their own battery; no external verification or appeal exists
yet; revision authority over the theoretical core lies outside the
authors' control while revision authority over the adaptation is
internal and so far undisclosed.

\item \textbf{Skill-acquisition efficiency.}
\emph{Exemplar:} Chollet's definition of intelligence as
skill-acquisition efficiency over novel tasks {[}3{]} and its
operationalization in ARC-AGI-2 and the ARC Prize program
{[}4{]}{[}5{]}. \emph{Operative content scored:} intelligence as the
efficiency of acquiring skill on tasks the system was not prepared for,
measured against human ease on the private evaluation sets.
\emph{Source/year:} {[}3{]} (2019); {[}4{]} (2025); {[}5{]} (2026).
\emph{Primary unit:} skill-acquisition efficiency. \emph{Hybrid:} no.
\emph{Scores:} C1 high, the benchmark fixes tasks, success conditions
and a private evaluation protocol independent parties can run; C2 high
by the rule's second route, transfer to unprepared conditions is the
explicit target; C3 partial, indifference to autonomous goal-setting is
recoverable from the construct but unstated; C4 high, the ordinary
human confronting a novel puzzle is specified and the comparison is
data efficiency; C5 partial, the version-1-to-version-2 rebuild was versioned,
prospective and publicly justified, facts component (v) of the C6 audit
records, but the rebuild materially changed difficulty and broke direct
score comparability across versions, degrading the cross-version
interpretability that C5 measures. \emph{C6 audit:} authorship is an identified individual and a
nonprofit prize structure; material interest attaches to the
benchmark's adoption and the prize ecosystem, disclosed; certification
runs through a public competition protocol; partial external
verification exists through open participation, though task design
remains internal; revisability is versioned and publicly argued.

\item \textbf{Deflationary/value-laden boundary position.}
\emph{Exemplar:} the Unsocial Intelligence analysis {[}7{]}, with the
pragmatic variant implied when laboratory executives publicly question
the term's usefulness while invoking it. \emph{Operative content:} the
position that no single shared definition is available or desirable
because definitions of intelligence encode contested values.
\emph{Source/year:} {[}7{]} (2024). \emph{Primary unit:} refused.
\emph{Hybrid:} not applicable. \emph{Scores:} C1 low by design, the
position supplies no failure condition; C2 through C5 not defined, no
threshold exists whose breadth, explicitness, reference class or
stability could be assessed. \emph{C6 audit:} authorship is academic
and disclosed; no financial or contractual interest attaches to an
arrival verdict, while intellectual commitment and professional
interest in the critique's uptake exist and are recorded under the same
taxonomy applied to every other entry; there is no certification
because there is nothing to certify; the
position functions as a standing external critique of every other
entry's governance rather than as a candidate standard.

\end{enumerate}

The hybrid flag and the primary-unit assignments above implement the
selection rule of Section 3: where a definition couples units, the
family of its primary unit governs its row and the coupling is recorded
rather than suppressed, because Section 6 argues that exactly such
couplings carry the persuasive force of arrival claims.

\begin{center}\rule{0.5\linewidth}{0.5pt}\end{center}

\section*{Acknowledgments}

The author thanks the Universidad Internacional de Investigación México for institutional support of the postdoctoral research that underwrites the construct of definitional sovereignty introduced here and the readers whose feedback on earlier drafts sharpened the artifact and its limits.

\section*{References}
\small
\sloppy
\setlength{\parskip}{0.5em}
\begingroup
\setlength{\leftskip}{1.5em}
\setlength{\parindent}{-1.5em}

{[}1{]} Morris, M. R., Sohl-Dickstein, J., Fiedel, N., Warkentin, T.,
Dafoe, A., Faust, A., Farabet, C., \& Legg, S. (2024). Levels of AGI for
operationalizing progress on the path to AGI. \emph{Proceedings of the
41st International Conference on Machine Learning (ICML)}, PMLR 235,
36308--36321. arXiv:2311.02462. https://arxiv.org/abs/2311.02462

{[}2{]} Hendrycks, D., Song, D., Szegedy, C., Lee, H., Gal, Y.,
Brynjolfsson, E., Li, S., Marcus, G., Tegmark, M., Bengio, Y., et
al.~(2025). A definition of AGI. arXiv:2510.18212.
https://arxiv.org/abs/2510.18212 ; project site
https://www.agidefinition.ai/

{[}3{]} Chollet, F. (2019). On the measure of intelligence.
arXiv:1911.01547. https://arxiv.org/abs/1911.01547

{[}4{]} Chollet, F., Knoop, M., Kamradt, G., \& Landers, B. (2025).
ARC-AGI-2: A new challenge for frontier AI reasoning systems.
arXiv:2505.11831. https://arxiv.org/abs/2505.11831

{[}5{]} Chollet, F., Knoop, M., Kamradt, G., \& Landers, B. (2026). ARC
Prize 2025: Technical report. arXiv:2601.10904.
https://arxiv.org/abs/2601.10904

{[}6{]} Grace, K., Stewart, H., Sandkühler, J. F., Thomas, S.,
Weinstein-Raun, B., \& Brauner, J. (2024). Thousands of AI authors on
the future of AI. arXiv:2401.02843.
https://doi.org/10.48550/arXiv.2401.02843

{[}7{]} Blili-Hamelin, B., Hancox-Li, L., \& Smart, A. (2024). Unsocial
intelligence: An investigation of the assumptions of AGI discourse.
\emph{Proceedings of the AAAI/ACM Conference on AI, Ethics and Society},
7(1), 141--155. https://doi.org/10.1609/aies.v7i1.31625 ; arXiv:2401.13142.
Developing the value-laden critique of intelligence measurement in
Blili-Hamelin, B., \& Hancox-Li, L. (2023), \emph{Making intelligence:
Ethical values in IQ and ML benchmarks}, ACM FAccT 2023.

{[}8{]} March, S. T., \& Smith, G. F. (1995). Design and natural science
research on information technology. \emph{Decision Support Systems},
15(4), 251--266.

{[}9{]} Hevner, A. R., March, S. T., Park, J., \& Ram, S. (2004).
Design science in information systems research. \emph{MIS Quarterly},
28(1), 75--105.

{[}10{]} Peffers, K., Tuunanen, T., Rothenberger, M. A., \& Chatterjee,
S. (2007). A design science research methodology for information systems
research. \emph{Journal of Management Information Systems}, 24(3),
45--77.

{[}11{]} Carnap, R. (1950). \emph{Logical foundations of probability}.
University of Chicago Press. (Notion of explication.)

{[}12{]} Chalmers, D. J. (2025). What is conceptual engineering and what
should it be? \emph{Inquiry: An Interdisciplinary Journal of Philosophy},
68(9), 2902--2919. https://doi.org/10.1080/0020174X.2020.1817141 (First
published online 2020.)

{[}13{]} Cappelen, H. (2018). \emph{Fixing language: An essay on
conceptual engineering}. Oxford University Press.

{[}14{]} Aguilera Briones, J. E. (2026). \emph{Sociedad Algorítmica
Autónoma en los Estados Unidos Mexicanos} {[}Postdoctoral research{]}.
Zenodo. https://doi.org/10.5281/zenodo.19232168 (Constructs of
algorithmic sovereignty, critical dependency and algorithmic
governance.)

{[}15{]} OpenAI (2018). \emph{OpenAI charter}. (AGI as highly autonomous
systems that outperform humans at most economically valuable work.)

{[}16{]} Sarma, G. P., Bhatt, S. D., Jacob, M., \& Steratore, R. (2025).
\emph{AGI forecasting} {[}Research Report RR-A4692-1{]}. RAND
Corporation.

{[}17{]} Schneider, W. J., \& McGrew, K. S. (2018). The
Cattell--Horn--Carroll theory of cognitive abilities. In D. P. Flanagan
\& E. M. McDonough (Eds.), \emph{Contemporary Intellectual Assessment:
Theories, Tests and Issues} (4th ed., pp.~73--163). Guilford Press.

{[}18{]} Venable, J., Pries-Heje, J., \& Baskerville, R. (2016). FEDS: A
framework for evaluation in design science research. \emph{European
Journal of Information Systems}, 25(1), 77--89.

{[}19{]} Gregor, S., \& Hevner, A. R. (2013). Positioning and presenting
design science research for maximum impact. \emph{MIS Quarterly}, 37(2),
337--355.

{[}20{]} Turing, A. M. (1950). Computing machinery and intelligence.
\emph{Mind}, 59(236), 433--460.

{[}21{]} Bubeck, S., Chandrasekaran, V., Eldan, R., Gehrke, J., Horvitz,
E., Kamar, E., Lee, P., Lee, Y. T., Li, Y., Lundberg, S., Nori, H.,
Palangi, H., Ribeiro, M. T., \& Zhang, Y. (2023). Sparks of artificial
general intelligence: Early experiments with GPT-4. arXiv:2303.12712.

{[}22{]} Legg, S., \& Hutter, M. (2007). A collection of definitions of
intelligence. In B. Goertzel \& P. Wang (Eds.), \emph{Advances in
Artificial General Intelligence} (Frontiers in Artificial Intelligence
and Applications, Vol. 157, pp.~17--24). IOS Press.

{[}23{]} Gallie, W. B. (1956). Essentially contested concepts.
\emph{Proceedings of the Aristotelian Society}, 56, 167--198.

{[}24{]} Espeland, W. N., \& Stevens, M. L. (1998). Commensuration as a
social process. \emph{Annual Review of Sociology}, 24, 313--343.

{[}25{]} Porter, T. M. (1995). \emph{Trust in numbers: The pursuit of
objectivity in science and public life}. Princeton University Press.

{[}26{]} Espeland, W. N., \& Sauder, M. (2007). Rankings and reactivity:
How public measures recreate social worlds. \emph{American Journal of
Sociology}, 113(1), 1--40.

{[}27{]} Soto, Á., Durán, R., Moreno, A., Adasme, S., Rovira, S.,
Jordán, V., \& Poveda, L. (Coords.). (2025). \emph{Índice Latinoamericano
de Inteligencia Artificial (ILIA) 2025} (Documentos de Proyectos
LC/TS.2025/68/Rev.1). Comisión Económica para América Latina y el Caribe
(CEPAL) \& Centro Nacional de Inteligencia Artificial (CENIA).

{[}28{]} Stanford Institute for Human-Centered Artificial Intelligence
(HAI). (2025). \emph{Artificial intelligence index report 2025}.
Stanford University.

{[}29{]} Microsoft (2025, October 28). \emph{The next chapter of
the Microsoft--OpenAI partnership}. Official Microsoft Blog.
\url{https://blogs.microsoft.com/blog/2025/10/28/the-next-chapter-of-the-microsoft-openai-partnership/}

{[}30{]} Anthropic (2026, February 24). \emph{Responsible Scaling
Policy, Version 3.0}. \url{https://www.anthropic.com/responsible-scaling-policy/rsp-v3-0}

{[}31{]} OpenAI (2025, April 15). \emph{Preparedness Framework,
Version 2}. \url{https://openai.com/index/updating-our-preparedness-framework/}

{[}32{]} Microsoft (2026, April 27). \emph{The next phase of the
Microsoft--OpenAI partnership}. Official Microsoft Blog.
\url{https://blogs.microsoft.com/blog/2026/04/27/the-next-phase-of-the-microsoft-openai-partnership/}

{[}33{]} Bowker, G. C., \& Star, S. L. (1999). \emph{Sorting things
out: Classification and its consequences}. MIT Press.

{[}34{]} Jasanoff, S. (Ed.). (2004). \emph{States of knowledge: The
co-production of science and the social order}. Routledge.

{[}35{]} B\"uthe, T., \& Mattli, W. (2011). \emph{The new global
rulers: The privatization of regulation in the world economy}.
Princeton University Press.

\endgroup

\end{document}